Same Input, Different Scores: A Multi Model Study on the Inconsistency of LLM Judge

**Fiona Lau**
Fiona.lau@hotmail.co.uk

**Abstract**

Large language models (LLMs) are increasingly used as automated evaluators in both research and enterprise settings, a practice often referred to as LLM-as-a-judge. While prior work has examined their accuracy, bias, and alignment with human preferences, far less is known about the consistency of LLMs when assigning numerical scores—particularly in score-based judgment tasks that underpin many production workflows. This study provides a systematic evaluation of scoring stability across five widely used models (GPT-4o, GPT-4o-mini, Gemini-2.5-Flash, Claude-Haiku-4.5, and Claude-Sonnet-4.5), two temperature settings, and real enterprise question–answer pairs sourced from a retrieval-augmented generation (RAG) system.

We investigate three research questions: (1) the consistency of a model's scoring across repeated runs, (2) cross-model differences in scoring identical inputs, and (3) the extent to which temperature influences scoring stability. Temperature is a decoding parameter that controls how deterministic or random an LLM's output is during inference. Our results show substantial inconsistency across all models, even under temperature=0 conditions, with the completeness criterion exhibiting the highest variability. Cross-model comparisons reveal systematic differences in strictness and interpretation across model families, leading to divergent ratings of the same responses. Temperature reductions improve stability for some models—most notably GPT-4o and Gemini—but have limited or inconsistent effects for Anthropic models.

These findings have significant implications for enterprise pipelines that rely on LLM-generated scores for routing, triage, gating logic, or quality checks. We show that identical inputs can receive different scores depending on model, model family and temperature, raising concerns around fairness, repeatability, and operational reliability. Our study underscores the need for careful monitoring, robust parsing strategies, and hybrid human-LLM evaluation approaches to ensure dependable use of LLM-as-a-judge in production environments.

## 1. Introduction

Since the public release of ChatGPT in 2022, the use of LLM has expanded rapidly. Beyond their early applications in content generation, translation, and summarization, LLMs are now widely used as evaluators—an approach commonly referred to as LLM-as-a-judge. In this setting, an LLM is tasked to assess the outputs of other LLMs or user-generated content [11][20], across many aspects, including reliability, relevance, and helpfulness [8]. Point-wise and pair/list wise comparison can both be done [8][27], the outputs of assessment could be score, ranking, or selection [8].

LLM-as-a-judge has become more popular due to its speed, scalability, flexibility [8], and explainability [27]. As systems using LLM can generate hundreds and thousands of output in minutes, it becomes very costly and time-consuming to rely entirely on human annotators, hence the LLM-as-a-judge has become an attractive alternative for large-scale evaluation. At the same time, traditional Natural Language Processing (NLP) evaluation methods, such as BLEU [13] and ROUGE [10] are not the most suitable method to measure open-ended and dynamic text output, as these methods only measure the lexical overlap between the output and the golden template [12]. LLM-as-a-judge is a more adaptive way of assessing the quality of generated content.

Recently researchers have studied the different aspects of LLM-as-a-judge, such as its biases [2][4][7][28], accuracy [27], safety [16], structure [17], and level of alignment with human preference [8][12][26]. However, one area remains comparatively understudied: the consistency of LLMs when acting as judges.

Prior work [1][14][19] shows that LLM shows low consistency in their judgement; the LLM generates a different output with the same prompt, for example, LLM would choose different answer for the same multiple-choice question when run repeatedly or generate different codes each time for the same prompt. The inconsistency is mainly due to the non-deterministic nature of LLM, as most LLMs use a SoftMax function at the top of its decoder to convert its internal representation into predicted next-token probabilities. The SoftMax layer indicates the probability of different words, and the final output depends on the strategy used during inference [22].

The transformer selects the next token based on the probabilistic distribution [22]. The process can be controlled by the parameters: temperature and top-k. Temperature setting is a technique used to shape the probability distribution by applying a parameter t to the raw scores before the final softmax calculation. The probability distribution is re-estimated by dividing the logits by the temperature t and exponentiating time, which means that setting the probability below 1.0 skews the distribution towards high-probability events, and high temperature can be used to maintain diversity, hence the creativity part of the LLM [3], while Top-k sampling is a method where the model samples the next word only from top k most probable choice [3]. Generated output can be affected by the length of prompts [7], the length of generated output [2], the penalty score [1], and the sampling method [5][21]. Although in theory, having the same prompt and setting temperature equals zero should result in 'deterministic output" [1], in practice, the single output from an LLM is a representation of a single draw from the model's distribution [20], therefore it is still random [14].

A smaller subset of research has specifically explored the stability of LLM decisions, such as answer consistency in multiple-choice tasks [20] or selection-based evaluations. However, these studies do not address an increasingly common evaluation setting: score-based judgment, where models must assign numerical values to qualitative criteria, for example, providing a score between 0 and 10 indicating how relevant the answer is to the question. Despite scoring being one of the most prevalent outputs used in enterprise evaluation pipelines, little is known about how stable these scores are across repeated trials or across different LLM families.

This study aims to fill this gap. We investigate the stability of score-based LLM judgements across multiple models, question types, and temperature settings. Our analysis addresses three research questions:

RQ1: How consistent is the model scoring one question and across different questions?
RQ2: How differently do various models score the same question?
RQ3: To what extent does temperature affect scoring consistency?

As LLM-based scoring becomes more deeply integrated into enterprise decision-making, to power workflows such as routing, triage, gating decisions, and quality checks, therefore understanding the limits of LLM consistency is essential. This work provides an empirical foundation for that understanding and highlights the risks and considerations involved in deploying score-based evaluators in production systems.

2. **Related work**

**LLM-as-a-judge**
The concept of LLM-as-a-judge was proposed by researchers [23][27] following the advancement of large language models (GPT-4 and GPT o1). LLM-as-judge can be implemented by assessing one output or comparing 2 or more outputs (pair/list-wise judgement) [27].
The most widely adopted protocol is the score-based judgement, in which the output could be a continuous or discrete score for quantitative comparison [9] or attribute detection [25]. Other forms of output include ranking and selection.
LLM-as-a-judge can provide judgement on various perspective, such as helpfulness, reliability, and relevance. LLMs assess the qualitative aspect of the generated output, which is hard to assess using traditional NLP method, such as BLEU [13] and ROUGE [10]. These perspectives provide a guidance on the performance of the LLM being judged.

**Non-deterministic nature of LLM**
The most relevant work on the study from Atil, which addresses the issue of instability of LLMs, i.e. the same prompt would produce different output, even under the supposed "deterministic" setting. In the research, they investigated LLMs non-determinism by setting the temperature at zero, top-p at 1, and fixing the seed. By running the LLMs across eight tasks in 20 runs, they look at the minimum and maximum accuracies. However, they focus on the accuracies of the tasks, and these are not score-based problems.

Meanwhile, another research focused on internal consistency across replicated evaluations [20], they investigated the "fixed randomness" – the inherent randomness remaining even when parameters like temperature are set to zero. However, their methodology is based on a selection task, where the LLM is prompted to choose the "best" response from a set of five multiple-choice options.

Pinhanez's research examines the answer consistency of smaller LLMs (2B-8B parameters) [14], their research shows that while medium-sized models show higher level of consistency, small models typically are less consistent. Again, this work focuses on multiple-choice contexts only.
While research offer the methodologies for quantifying the non-determinism and reliability of LLMs, they primarily focus on selection-based or multiple-choice tasks. None of the studies address the instability of a score-based judging system, where the LLM is required to generate a numerical score to a given response.

### 3. Experiment

The questions used in this study were sourced from an enterprise chatbot deployment built on a retrieval-augmented generation (RAG) architecture. As these questions originate from real user interactions, they naturally span a diverse range of categories and reflect authentic information-seeking behaviour. Due to privacy constraints and the inclusion of proprietary organizational knowledge in the answers, the dataset cannot be released publicly. The use of an internal enterprise dataset also reduces the risk of benchmark contamination [18], whereby LLMs demonstrate inflated performance due to prior exposure to widely used public benchmarks during their training and fine-tuning process.

At the same time, it is important to show case how easily in a business setting, a misuse of model could create issues.

| Attribute | Definition |
| --- | --- |
| Relevance | whether the answer directly addresses the question |
| Reliability | whether the information provided is factually correct |
| Completeness | whether the answer fully covers the information needed to respond to the question |

Table 1 – Evaluation Criteria for Answer Assessment

While relevance and accuracy are common in automated evaluation, completeness is less frequently used. We include it because completeness provides insight into an essential aspect of RAG system performance: whether the retrieval component supplies the LLM with sufficient supporting information to generate a comprehensive response. Thus, completeness allows direct assessment of retrieval sufficiency, complementing accuracy and relevance.

All models used the same evaluation prompt [Appendix 1] and were instructed to assign scores between 0 and 1 for each criterion, accompanied by a textual justification.

To examine both cross-model and within-family differences, we selected three major families of LLMs widely used in enterprise environments: OpenAI (ChatGPT), Google (Gemini), and Anthropic (Claude).
From each family, two models were chosen, enabling comparison between models sharing similar architectural foundations. Each model was evaluated under two temperature settings, temperature = 0 and temperature = 1, to investigate how sampling stochasticity affects scoring consistency and explanation quality.

The five models included in the study are GPT-4o, GPT-4o-mini, Gemini-2.5-Flash, Claude-Haiku-4.5, Claude-Sonnet-4.5.

## 4. Results

RQ1: How consistent is the model scoring one question and across questions?

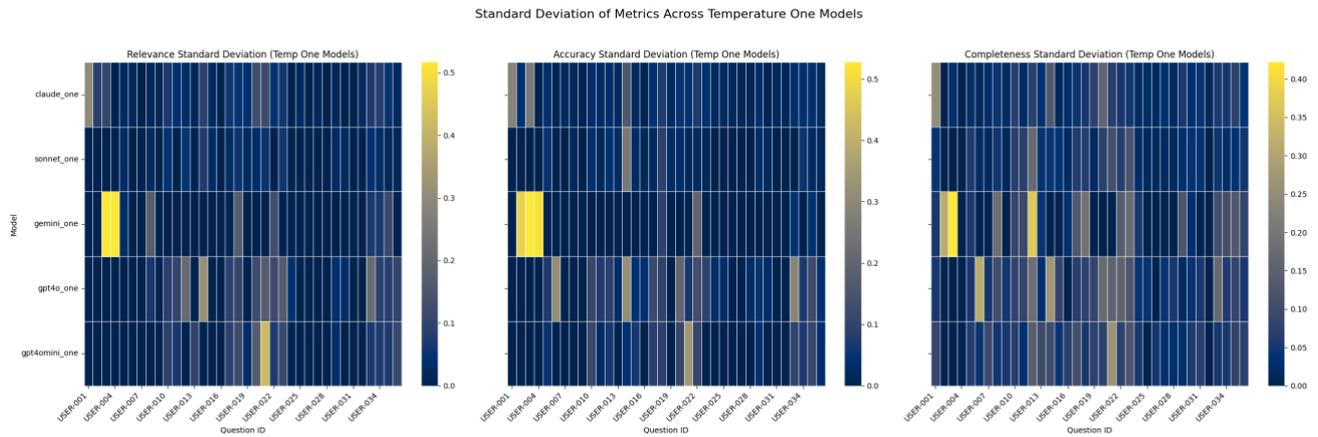

*Figure 1: Score variability across models (all questions) (Temperature=1)*

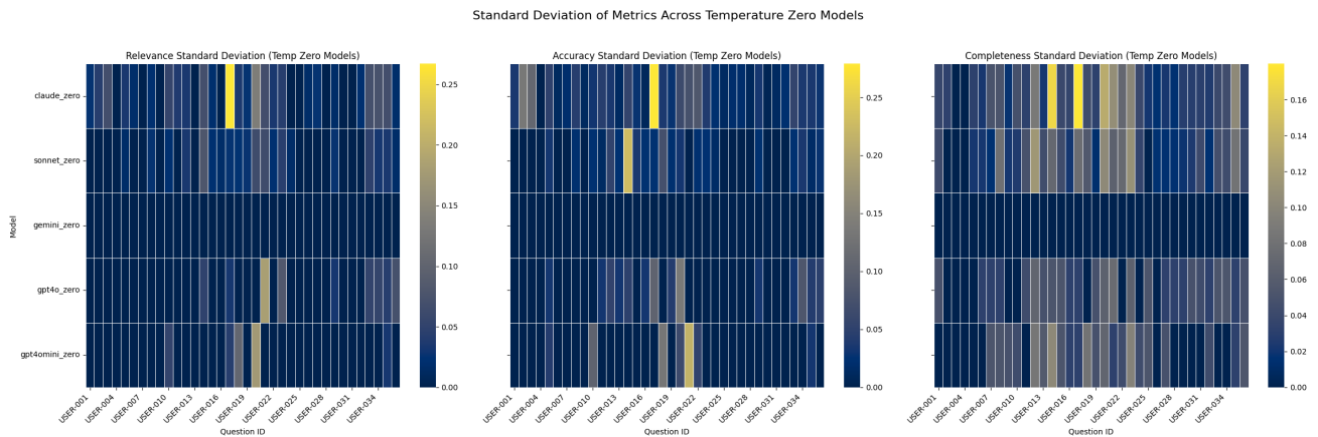

*Figure 2: Score variability across models (all questions) (Temperature = 0)*

To evaluate within-model consistency, we examined how stably each model assigned scores when presented with the same question over repeated runs. Figures 1 and 2 illustrate the standard deviation of scores across ten runs for each model at temperature = 1 and temperature = 0, respectively. At temperature 0, models do not exhibit deterministic behaviour in practice. Despite the expectation of stability under a zero-temperature setting, several models generated noticeably variable scores across repeated trials. This is especially evident in the prominent yellow spikes in Claude Zero, and the patchy yellow regions in Sonnet Zero and GPT-4o Mini Zero (Figure 2). These patterns indicate that the same model, under nominally deterministic sampling, still assigns different numeric ratings to identical prompts.

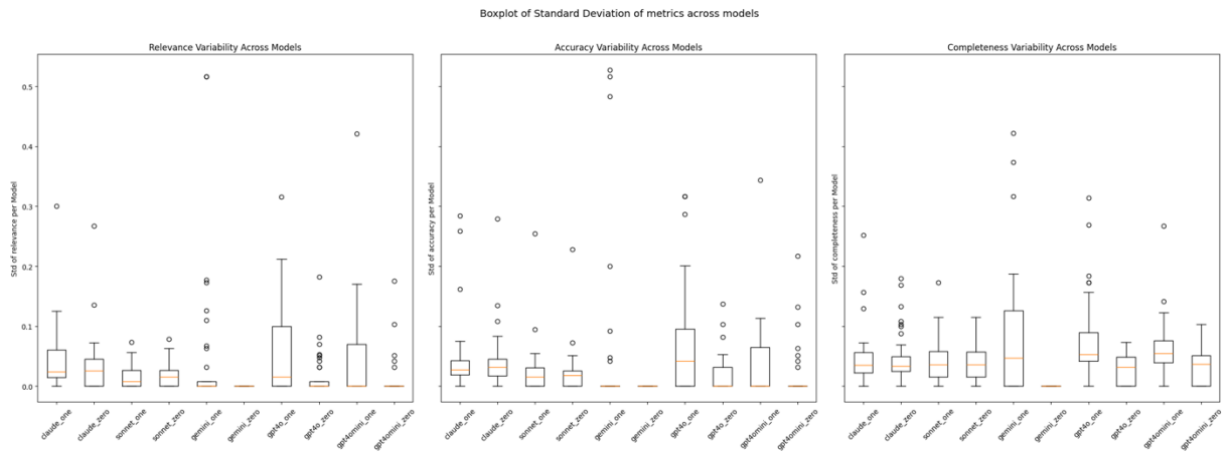

*Figure 3: Standard Deviation of Evaluation Metrics Across Models (Boxplot)*

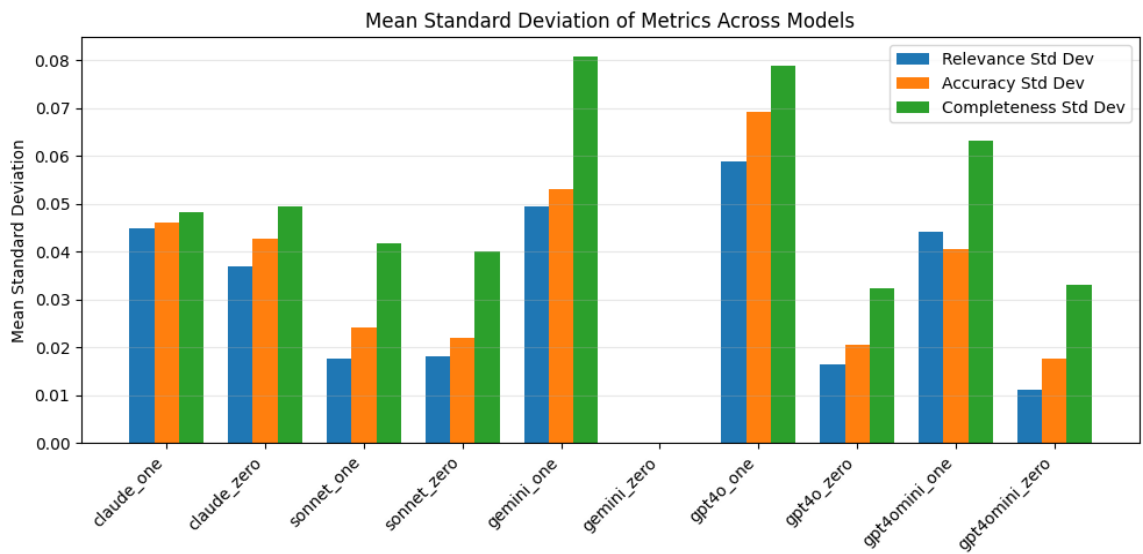

*Figure 4: Mean Standard deviation of evaluation metrics across models*

Models rate more consistently in relevance and accuracy than in completeness, there are less yellow light patches in the relevance map, compare accuracy and completeness. This is also supported by the bigger boxplot in completeness in figure 3, as well as in figure 4, which relevance has the lowest standard deviation across most models, except GPT4o-mini One, where relevance has a higher standard deviation than accuracy. Also notice that most temperature=zero version has a lower standard deviation than the temperature = 1 version of models, however, the reduction depends on the models. The anthropic family (Claude and Sonnet), the reduction in standard deviation is minimal, noted that the completeness standard deviation of Claude zero is higher than the Claude one.

However, there are noticeable differences in the GPT family, both GPT4o and GPT4o-mini has a reduction in standard deviation. The most noticeable is the Gemini model, where standard deviation is dropped to zero.

From figure 3, we also notice that although the mean standard deviation remains low, there are a lot of outliers in the graph. For accuracy, all models except Sonnet models, have multiple outliers on the boxplots, the most extreme standard deviation comes from Gemini one. For accuracy, most models have more outliers, and the anthropic models have a very similar plot between Claude (1/0) and Sonnet (1/0). The GPT4o models has the thickest boxplot among all. For completeness, all the boxplots are bigger than the previous, which means there are more variations in the scoring, which is also supported by figure 1 and 2.

Figure 3 shows that Gemini One has a higher consistency across all questions in metric accuracy and relevance. however, in figure 1 & 2, we notice that in a few questions (USER-002 to USER-004), it demonstrates high inconsistency. Figure 5 captures this phenomenon, Gemini rate the question either 1 or 0, across multiple runs, hence the high inconsistency. However, for the other questions, the model shows high consistency, rating most of the questions 1, across multiple runs. Only in a few questions, the rating varies [figure 5]

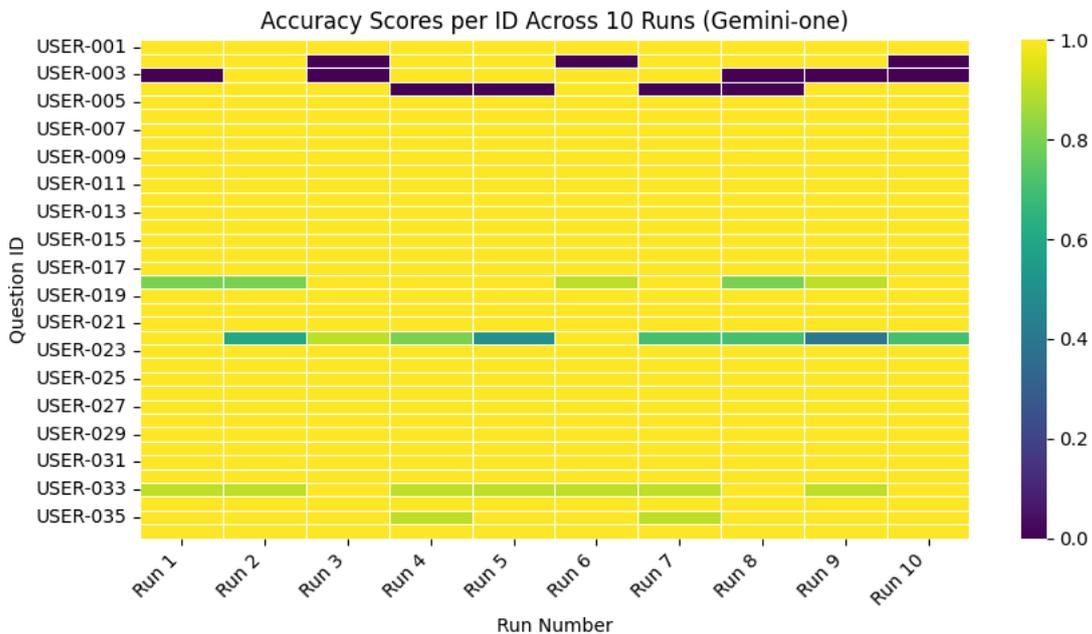

*Figure 5: Accuracy score per question across 10 runs (Gemini-One)*

For the Anthropic models, the Claude and Sonnet models are more consistent at Temperature = 1, notice the yellow spikes (high inconsistency) in Figure 2 from Claude and Sonnet for temperature 0, especially in the completeness graph. Figure 6 shows the individual scoring for each question, and how varied the scores are.

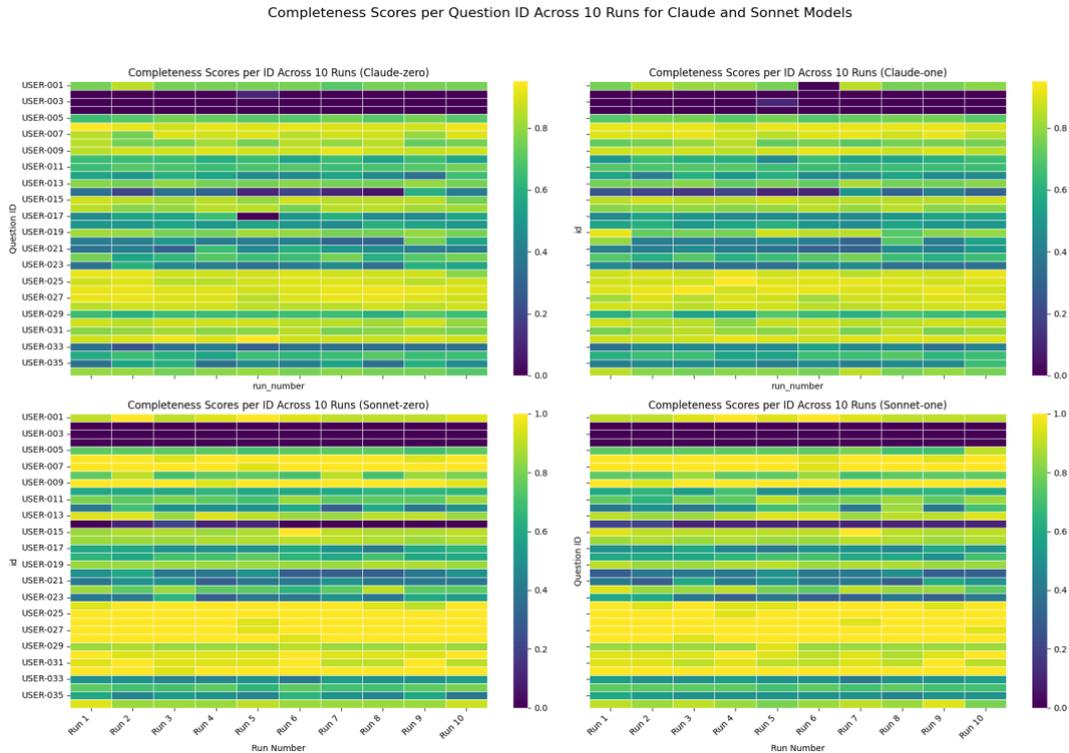

*Figure 6: Completeness Scores for 10 runs (Anthropic models)*

In summary, models demonstrate significant inconsistency when scoring identical questions across multiple runs—particularly for the completeness metric, and especially under higher temperature settings. Even at temperature 0, none of the evaluated models achieve full determinism, and the degree of variability varies significantly across model families.

RQ2: How differently do various models score the same question?

To assess cross-model variation, we compared how each model assigned scores to the same set of questions across relevance, accuracy, and completeness. The mean-score heatmaps (Figures 7–9) reveal substantial systematic differences across model families.

Completeness exhibits the strongest cross-model divergence. While most models give lower completeness scores than they do for relevance or accuracy, the Gemini models again stand out as the most generous, assigning higher completeness ratings across questions compared with GPT and Anthropic models.

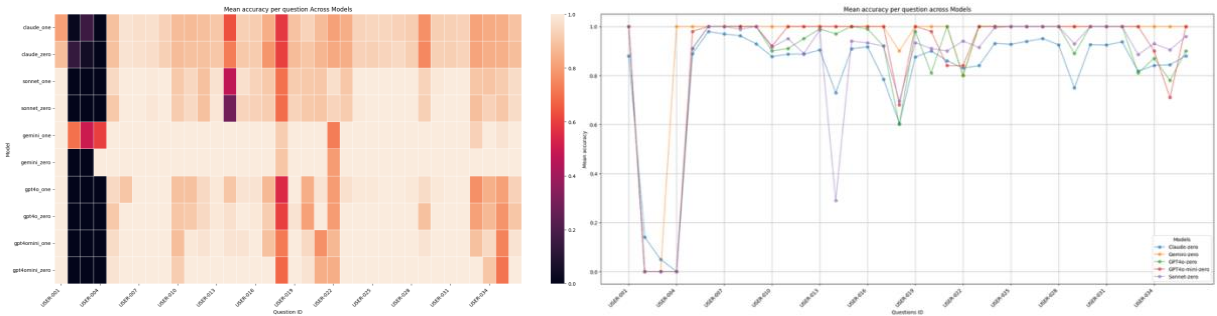

*Figure 7: Mean Accuracy scores across models*

Question-level inconsistencies highlight these differences further. For Question 14, both relevance and accuracy show pronounced disagreement across model families. While most models rate the answer as highly relevant and accurate, both Claude models assign near-zero scores, interpreting the response as fundamentally misaligned with the user's intent. In their view, the answer fails because it describes business tools rather than the assistant's own capabilities—an interpretation not shared by GPT or Gemini models, which rate the answer positively.

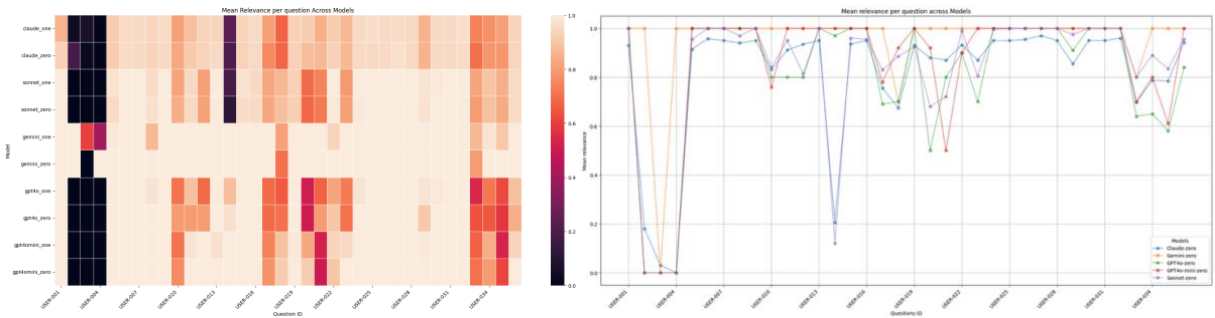

*Figure 8: Mean Relevance score across models*

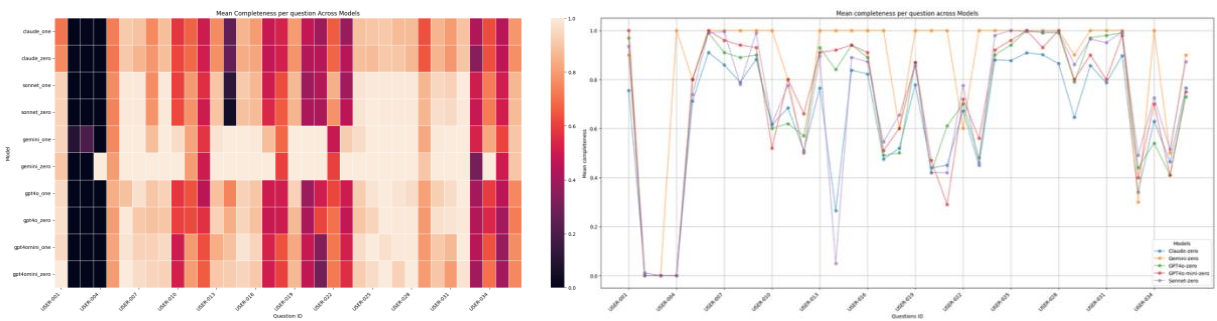

*Figure 9: Mean Completeness score across models*

Similarly, Question 21 illustrates substantial divergence in completeness. The Gemini models assign a completeness score of 1, arguing that the answer fully explains the inability to provide the requested sequence diagram and clearly justifies the limitation. In contrast, GPT-4 evaluates the same answer as only partially complete (0.5), noting that although it explains the lack of information, it does not provide follow-up suggestions or alternative resources, and thus falls short of a fully complete response.

These examples demonstrate that model families differ not only in strictness or generosity but also in their interpretation of what constitutes relevance, accuracy, and completeness. Such disparities have meaningful implications for business workflows: a single question could be classified differently depending on the model used, potentially triggering inconsistent downstream actions, such as routing decisions, gating logic, or customer-facing outcomes.

RQ3: To what extent does temperature affect scoring consistency?

Temperature affects consistency unevenly across model families and evaluation dimensions. Most models show reduced variance at temperature 0 compared to temperature 1, but the magnitude of improvement varies.

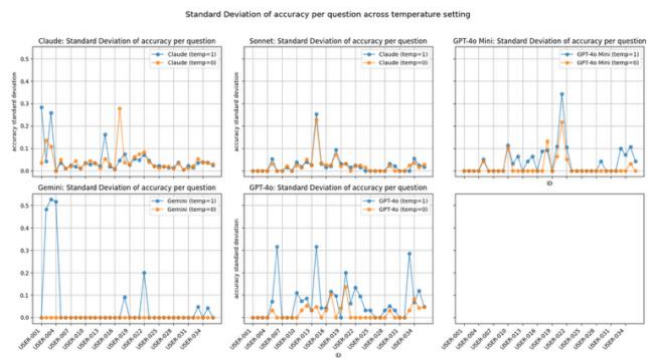

*Figure 10: Standard Deviation of accuracy per question across temperature settings*

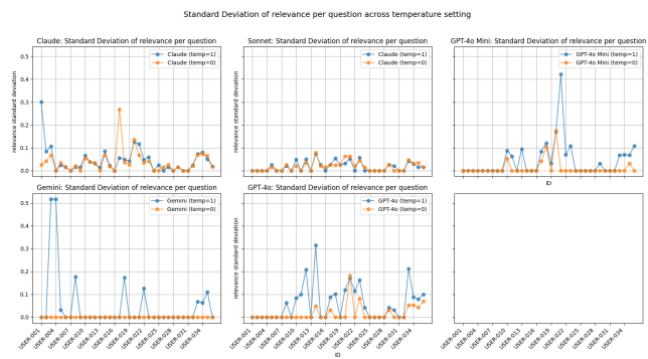

*Figure 11: Standard Deviation of relevance per question across temperature settings*

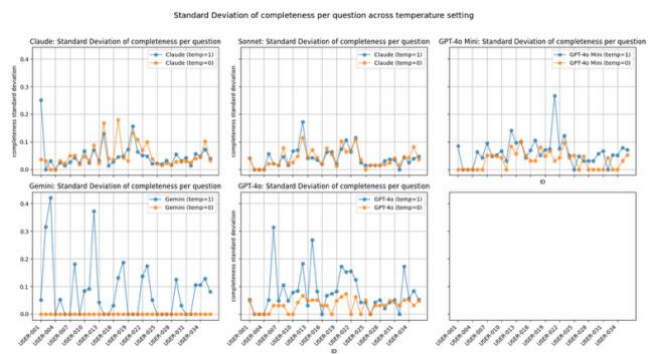

*Figure 12: Standard Deviation of completeness per question across temperature settings*

Figure 10 -12 shows that Anthropic models (Claude and Sonnet) show only marginal reductions in variance, and in some cases (e.g., Claude Zero on completeness), variability remains high or even increases.

By comparison, GPT models (GPT-4o and GPT-4o-mini) exhibit clearer reductions in standard deviation when temperature is lowered, suggesting that their scoring behaviour becomes more stable when stochasticity is minimized.

Gemini model displays the strongest stabilising effect of temperature: at temperature 0, the variance in several cases drops to nearly zero, indicating highly consistent behaviour across runs relative to other model families.

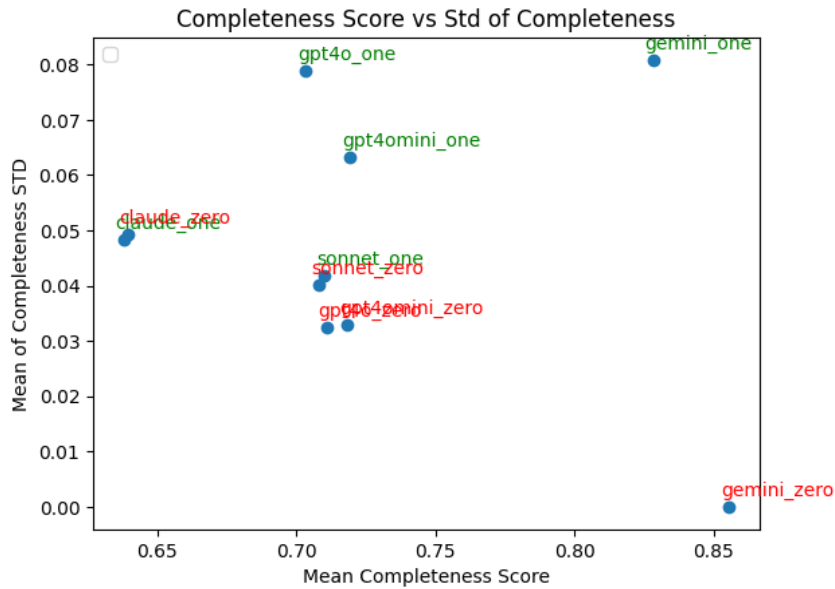

*Figure 13: Completeness Score vs Std of Completeness*

Figure 13 further illustrates the relationship between mean completeness scores and their corresponding standard deviations. Notably, setting temperature to zero does not guarantee substantial reductions in instability: the overall level of variation across questions remains similar for most models, particularly within the Anthropic family. The Gemini models again represent the exception, showing a near-complete reduction of variability under the zero-temperature condition. Additionally, there is no clear correlation between a model's average completeness score and its standard deviation—models can exhibit similar levels of variability even when their mean scores differ substantially.

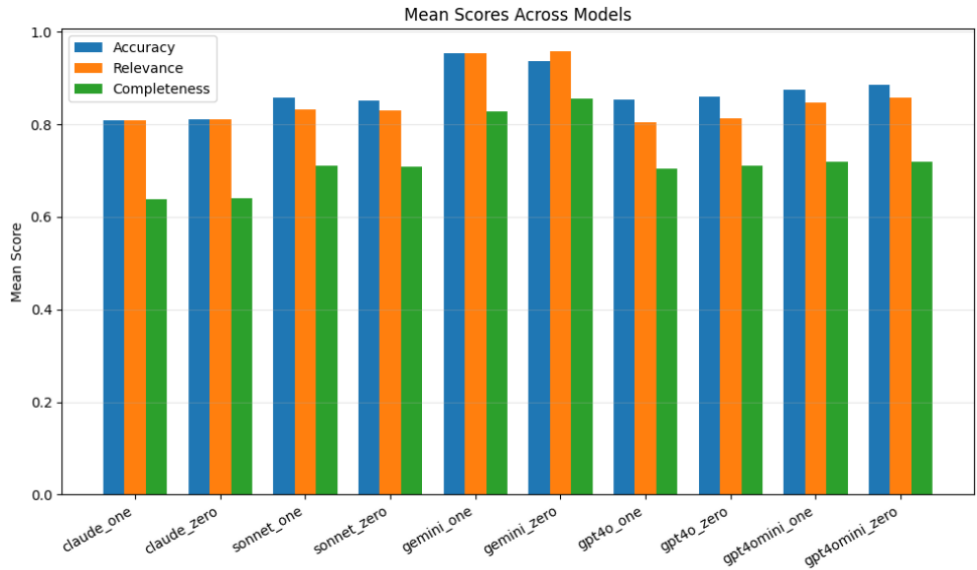

*Figure 14: Mean score of metrics across models*

Figure 14 shows that the mean scores across temperature settings remain broadly consistent, indicating that temperature affects stability rather than the direction or tendency of scoring.

Overall, increasing temperature consistently amplifies scoring randomness, with the strongest effects observed in models already prone to fluctuation, such as GPT-4o and Gemini at higher temperatures. Lowering temperature improves consistency, but only reliably for certain model families, underscoring that temperature-controlled determinism is far from uniform across LLM architectures.

## 5. Discussion

Our findings demonstrate that large language models exhibit substantial inconsistency when acting as evaluators—even when scoring identical question–answer pairs under controlled conditions. Despite using fixed prompts and identical inputs, models frequently produced different scores across repeated runs, underscoring the practical challenges posed by the inherent non-determinism of LLMs. This instability is especially pronounced for the completeness metric, where variance was consistently higher across models and temperature settings. Given that completeness is critical for evaluating retrieval-augmented systems, this unreliability presents a significant concern for enterprise use cases.

The implications of scoring variability extend directly into workflow risk. In many real-world deployments, LLM-based scoring is used as a gatekeeping mechanism for downstream actions, such as routing customer queries, triaging complaints, or determining follow-up steps in automated pipelines. When identical inputs can lead to different scores—as seen in cases like Question 21, where some models graded the response as fully complete while others assigned partial credit—customers may be treated inconsistently, raising concerns about fairness and potentially exposing organisations to legal implications.

While temperature adjustments can reduce inconsistency, our results show that this mitigation strategy is far from reliable. The degree of improvement varies significantly across model families: GPT and Gemini models benefit more noticeably from lower temperatures, whereas Anthropic models often show minimal reductions and, in some cases, increased variability at temperature 0. This suggests that non-determinism is not solely a function of sampling stochasticity but may reflect deeper architectural or decoding-level differences.

Another notable issue observed during the experiment is formatting inconsistency. Even when models assigned similar scores, their output formats occasionally deviated from the required structure. Such deviations caused downstream parsing errors, which in turn led to miscalculated or invalid scores. As a result, relying solely on structural assumptions or simple pattern matching (e.g., regular expressions) is fragile. A more robust approach—such as employing a secondary parser model or a dedicated extraction agent—is advisable to ensure evaluators produce well-formed and machine-readable outputs consistently.

Taken together, these findings highlight the importance of monitoring scoring consistency both within a single model and across model families. As LLMs are increasingly integrated into enterprise workflows—particularly in systems where automated judgement influences user experience, content classification, or operational decisions—it becomes essential to quantify and mitigate the risks associated with inconsistent evaluation. Understanding each model's variability profile enables more informed model selection, safeguards downstream fairness, and helps organisations design more reliable evaluation pipelines.

## 6. Future Work & Limitations

This study opens several avenues for further exploration. One natural extension is to examine how prompt design influences scoring stability, and whether prompt-engineering strategies can meaningfully reduce inconsistency across runs. Given that our evaluation relied on a single, fixed prompt template, future work could test alternative formulations—such as more structured instructions, chain-of-thought rationales, or explicit constraints—to determine whether certain prompting styles mitigate variability.

Another direction is to broaden the scope of evaluated models. Due to resource constraints, we tested only a subset of widely used proprietary LLMs. Expanding the study to include open-source models, domain-specialized models, or models of different parameter scales would provide a more comprehensive understanding of non-determinism in LLM-as-a-judge settings.

A further opportunity lies in combining automated evaluation with human expert judgment. Comparing model-generated scores against human-annotated baselines would help determine which models align more closely with human preferences and where discrepancies are most pronounced. Such analysis would also shed light on whether inconsistency correlates with a lack of human-aligned reasoning or whether the instability is orthogonal to human agreement patterns.

This work also has limitations. Our dataset is drawn from an internal enterprise RAG system, meaning that questions reflect real user behaviour but cannot be publicly released due to privacy and proprietary constraints. While this improves ecological validity, it limits full reproducibility and prevents broader benchmarking.

Finally, although we assessed model behaviour across two temperature settings, other decoding parameters—such as top-p, top-k, or seed control—were not systematically explored. Investigating how these parameters interact with model architecture to influence scoring stability represents an important direction for future research.

## 7. Conclusion

This study provides a systematic examination of the consistency of large language models when used as automated evaluators in score-based judgment settings. Despite the widespread adoption of LLM-as-a-judge in both research and enterprise applications, our results show that current models exhibit substantial instability when assigning numerical ratings—even under controlled, repeatable conditions. Across multiple model families, temperature settings, and evaluation criteria, none of the models demonstrated fully deterministic behaviour. Variation was especially pronounced for the completeness metric, which is critical for assessing retrieval-augmented systems but consistently yielded the highest levels of fluctuation.

The three research questions explored complementary dimensions of this instability. RQ1 revealed that models frequently produce different scores for the same question across repeated runs, with temperature-0 settings failing to eliminate randomness in several architectures. RQ2 showed that models also diverge meaningfully from one another when scoring identical inputs, indicating that evaluation outcomes can depend heavily on the choice of model family. RQ3 demonstrated that while lowering temperature generally improves stability, the effect is uneven: GPT and Gemini models benefit more from deterministic sampling, whereas Anthropic models continue to exhibit substantial variability even at temperature 0.

These findings have direct implications for enterprise workflows, where LLM-based scoring often underpins automated decisions such as query routing, content triage, system gating, and customer-facing interactions.  If identical inputs can yield different scores across runs or across models, downstream actions may become inconsistent, raising concerns around reliability, fairness, and auditability. The non-deterministic nature and unpredictability of LLM also raise concerns in deploying LLM in other more sensitive settings in the society, such as legal claims [6] and safety-critical software maintenance [15]. The study also highlights operational challenges, such as formatting inconsistencies that can cause parsing failures—even when the underlying scores are similar.

Building a reliable business system requires a reliable LLM system. Reliability, in this context, comes from the consistency of an LLM's output [20]. A model can even be consistently wrong and still be considered reliable, because its judgments are stable. If it always rates a question as 1, users can safely disregard that assessment, knowing the model is consistently incorrect. However, if the same model sometimes rates a question as 1 and other times as 0, users have no way to determine which judgment to trust or discard, even if one of them is correct. For this reason, it is more beneficial for businesses to evaluate LLMs not only by their mean score, but also by their standard deviation or confidence level. These additional measures help organisations assess and monitor models not just by their average performance, but also by stability and consistency.

Overall, our work underscores the need for caution when deploying LLM-as-a-judge in production systems. Non-determinism is not merely a theoretical property of generative models but a practical source of instability that must be measured, monitored, and mitigated. Improving evaluation robustness may require combining deterministic decoding strategies, prompt design advances, secondary parsing mechanisms, and human oversight. As organisations increasingly integrate LLMs into evaluation pipelines, understanding the limits of model consistency is essential for building dependable, fair, and trustworthy AI systems. This study offers a foundation for that understanding and identifies several paths for future research toward more stable and reliable automated evaluation frameworks.

# Appendix 1

```
You are an expert evaluator for question answering systems.
Please evaluate the following answer to a question based on the criteria below.

Question: {question}
Answer: {answer}

Rate the answer on the following criteria on a scale from 0 to 1,
where 0 is the lowest and 1 is the highest:

Criteria:
- Relevance: Does the answer directly address the question that was asked?
- Accuracy: Is the information provided in the answer factually, correct?
- Completeness: Does the answer provide a complete response to the question?

If the answer says "I don't know" or "I am sorry, I cannot find any information"
or "I am sorry, I cannot answer this question" or something similar or is empty,
give a score of 0 for all criteria.

Your evaluation must be formatted exactly as follows:

Relevance:
[Your reasoning here]
Score: [score between 0 and 1]

Accuracy:
[Your reasoning here]
Score: [score between 0 and 1]

Completeness:
[Your reasoning here]
Score: [score between 0 and 1]

Finally, provide an overall assessment of the answer's quality.
```